\documentclass{article}

\usepackage{times}
\usepackage{graphicx} 
\usepackage{subfigure} 

\usepackage{natbib}

\usepackage{algorithm}
\usepackage{algorithmic}

\usepackage{hyperref}

\usepackage[accepted]{icml2014} 

\usepackage{tikz}
\usepackage{array}
\usepackage{color}
\usepackage{amssymb}
\usepackage{amsmath}
\usepackage{amsfonts}
\usetikzlibrary{fit,positioning}

\icmltitlerunning{Prediction of Weak Effects}

\begin{document} 

\twocolumn[
\icmltitle{Bayesian Information Sharing Between Noise And Regression Models Improves Prediction of Weak Effects}

\icmlauthor{Jussi Gillberg$^\mathrm{1}$}{jussi.gillberg@aalto.fi}
\icmlauthor{Pekka Marttinen$^\mathrm{1}$}{}
\icmlauthor{Matti Pirinen$^\mathrm{2}$}{}
\icmlauthor{Antti J Kangas$^\mathrm{3}$}{}
\icmlauthor{Pasi Soininen$^\mathrm{3,4}$}{}
\icmlauthor{Marjo-Riitta J\"arvelin$^\mathrm{5}$}{}
\icmlauthor{Mika Ala-Korpela$^\mathrm{3,4,6}$}{}
\icmlauthor{Samuel Kaski$^\mathrm{1,7}$}{samuel.kaski@aalto.fi}

\icmladdress{$\mathrm{1}$) Helsinki Institute for Information Technology HIIT, Department of Information and Computer Science, Aalto University}
\vspace{-2mm}
\icmladdress{$\mathrm{2}$) Institute for Molecular Medicine Finland (FIMM), University of Helsinki, Finland}
\vspace{-2mm}
\icmladdress{$\mathrm{3}$) Computational Medicine, Institute of Health Sciences, University of Oulu and Oulu University Hospital, Oulu, Finland}
\vspace{-2mm}
\icmladdress{$\mathrm{4}$) NMR Metabolomics Laboratory, School of Pharmacy, University of Eastern Finland, Kuopio, Finland}
\vspace{-2mm}
\icmladdress{$\mathrm{5}$) Department of Epidemiology and Biostatistics, MRC Health Protection Agency (HPA) Centre for Environment and Health, School of Public Health, Imperial College London, United Kingdom; 
Institute of Health Sciences, University of Oulu, Finland;
Biocenter Oulu, University of Oulu, Finland;
Unit of Primary Care, Oulu University Hospital, Oulu, Finland;
Department of Children and Young People and Families, National Institute for Health and Welfare, Oulu, Finland
}
\vspace{-2mm}
\icmladdress{$\mathrm{6}$) Computational Medicine, School of Social and Community Medicine and the Medical Research Council Integrative Epidemiology Unit, University of Bristol, Bristol, UK}

\vspace{-2mm}
\icmladdress{$\mathrm{7}$) Department of Computer Science, University of Helsinki}

\icmlkeywords{multiple-output regression, weak effects, infinite Bayesian shrinkage, sparsity}

\vskip 0.3in
]

\begin{abstract} 
We consider the prediction of weak effects in a multiple-output regression setup, when covariates are expected
to explain a small amount, less than $\approx 1\%$, of the variance of the target variables. To facilitate the
prediction of the weak effects, we constrain our model structure by introducing a novel Bayesian approach of
sharing information between the regression model and the noise model. Further reduction of the effective
number of parameters is achieved by introducing an infinite shrinkage prior and group sparsity in the context
of the Bayesian reduced rank regression, and using the Bayesian infinite factor model as a flexible low-rank
noise model. In our experiments the model incorporating the novelties outperformed alternatives in genomic
prediction of rich phenotype data. In particular, the information sharing between the noise and regression
models led to significant improvement in prediction accuracy.%
%
%
%
\end{abstract} 

\section{Introduction}

Weak effects are ubiquitous in applications with genomic data. For example, individual genetic single
nucleotide polymorphisms (SNPs) may explain at most $\approx 1\%$ of the variance of the phenotype (i.e., the
response variable). Earlier such data has mainly been used for finding associations, but recently a need to
predict weak effects has emerged in plant breeding (in the applications of genomic prediction and genomic
selection) and is currently appearing in medicine.

The fundamental requirement for the prediction of weak effects is that a large number of samples is available
and consequently the statistical methods used must be scalable to large sample sizes.
From the statistical point of view, the investigation of weak effects can be enhanced by imposing as much
structure on the model as possible to minimize the effective number of parameters. 
Various ways of imposing structure on a learning problem have been proposed in the machine learning literature
as partial solutions to the so-called ``small $n$, large $p$'' problem where the number of variables $p$ is
large compared to the available sample size $n$. When the effects are weak, similar techniques are needed even
with large $n$. This challenging problem seems to have attained much less attention, although this kind of
data sets are ever more abundant for example in the biomedical application fields. 
When predicting weak effects, other partially related data are often available and predictive power can be
boosted by taking advantage of such data sources. For instance, when combining genomic data to phenotype data,
multiple related phenotype measurements are often available and simultaneous learning with all the phenotypes
is potentially valuable. Also an ``intelligent'' noise model is needed in the context of weak effects to
explain away the effects of latent confounding factors \cite{Fusi12}, which often are orders of magnitude
larger than the interesting effects, and, consequently, may result in reduced power if not accounted for.
Finally, Bayesian inference helps in quantifying the uncertainty inherent in any statistical analysis
involving weak effects.

In this work, we address the problem of predicting weak effects and propose a new method that brings together
several principles that are required for satisfactory prediction performance. The recent successful machine
learning techniques of infinite shrinkage priors and group sparsity are for the first time introduced in the
context of Bayesian reduced rank regression. However, these alone do not provide sufficient reduction in the
effective number of parameters and our main technical contribution is the introduction of a conceptually new
principle of sharing information between the regression model and the noise models in a Bayesian
multiple-output regression framework to provide additional regularization.

We use the Bayesian reduced rank regression \cite{Geweke96} as the basis for our multiple-output regression
model, as it encapsulates the prior information of the architecture of the multivariate input-output
relationships in an intuitive way. The reduced rank regression is obtained by considering a low-rank
approximation to the full regression coefficient matrix. Two structural assumptions, plausible in many
application domains, regarding the expected effects, immediately follow from this construction: first, if a
predictor has an effect on some response variable, then the predictor is likely to have an effect on other
responses as well; second, if a response is affected by a predictor, then the response is likely to be
affected by other predictors as well. 

To explain away confounders \cite{Fusi12, rai2012}, we learn a low-rank approximation for the error covariance
matrix by using an infinite Bayesian sparse factor model \cite{bhattacharya2011sparse} as the noise model. We
derive a novel infinite shrinkage prior for the reduced rank regression weight matrix by building on
assumptions similar to those of the noise model and provide proofs establishing the soundness of the prior in
the regression context. The prior not only helps avoid the problem of choosing a fixed rank for the model, but
also fixes the rotational invariance of the matrix factorization by enforcing the most significant effects to
be modeled by the first components. Consequently, interpreting the model becomes easier as no artificial
constraints on the regression weight matrix to ensure identifiability are needed.

Group sparsity of the effects on response groups is assumed, so as to incorporate prior knowledge about target
variable data. Group sparsity is achieved by using group-specific shrinkage priors and this is novel in the
context of reduced rank regression. We note that here we implicitly assume any effect from predictors to the
responses to affect several responses simultaneously (unless singleton response groups are defined). Thus, any
sparse variation in the high-dimensional response data is considered noise, justifying our use of the infinite
sparse noise model, where sparsity is capable of accounting for sparse variation whereas infinite rank adds
flexibility to account for the global noise component.

The conceptually new principle of directly sharing information between the regression model and the noise
model provides a powerful way to further reduce the effective number of parameters. In particular, we
incorporate the assumption that the effects of a predictor on correlated response variables are also
correlated. While group sparsity encourages the \emph{scales} of weight parameters within correlated response
groups to be similar, the new assumption also enforces the \emph{directions} of the effects to be similar
within the groups. The emerging model is analogous to the regression models with a G-prior
\cite{zellner1986assessing} where the weights of correlated covariates are assumed to be correlated; similar
structure is now assumed on the response variables. In practice, this is achieved by defining a joint prior
distribution for the regression weight matrix and the covariance matrix.

In an experiment, we study the impact of the different principles in predicting weak effects and compare the
new models to other methods. Rich phenotypes consisting of metabolomics data with 96 traits are predicted
using SNP data from 125 genes known to be associated with the phenotypes. Our model, entitled here as the
\textit{information sharing Bayesian reduced rank regression} (sharing BRRR), outperforms alternative methods
in prediction accuracy. A scalable R-implementation is made available upon publication of this article. 

%

\section{Related work}

Integrating multiple real-valued prediction tasks with the same set of covariates is called multiple-output regression \cite{breiman1997predicting}. The data consists of $N$ input-output pairs $(\mathbf{x}_n, \mathbf{y}_n)_{n= 1  \ldots N}$ where $P$-dimensional input vectors $\mathbf{x}$ (covariates) are used to predict $K$-dimensional vectors $\mathbf{y}$ of target variables. Methods for this setup have been proposed in different fields. In the application fields of genomic selection and Multi-trait Quantitative Trait Loci mapping, 
solutions \cite{Yi09, Xu08, Calus11, stephens2013unified} for low-dimensional target variable vectors ($K <
10$) have been proposed, but these methods do not scale up to our current needs as we want to pool as many
related prediction tasks as possible. 

Multi-task learning \cite{caruana1997multitask, Baxter96} is a machine learning genre for sharing statistical
strength between related learning tasks. We compare our method with the multi-task regression method
implemented in \verb glmnet  ~package \cite{glmnet} that allows elastic net regularization. Effects of
different regularization penalities have recently been studied and we run a continuum of mixtures of L1 and L2
penalties ranging from group lasso to ridge regression penalty. These methods do not use a noise model to
explain away confounders.

Sparse multiple-output regression models have been proposed for prediction of phenotypes from genomic data. To
relate to these works, we compare the predictive performance of our method to that the Graph-guided Fused
Lasso  (GFlasso \verb gw2 ) presented in \citet{kim2009multivariate}. This line of work has been very
successful. As compared to GFlasso, our approach is fully Bayesian and among other benefits, the fully
Bayesian treatment does not require cross-validation of regularization parameters.

Factor regression modeling is conceptually well suited for the problem at hand. To relate to this line of
work, we compare our model with the Bayesian infinite factor regression model derived in
\cite{bhattacharya2011sparse}.

Many promising methods for multi-task learning have been proposed in the field of kernel methods
\cite{evgeniou2007multi}, but these
methods will not, however, scale up to data sets with tens of thousands of samples, which are required as the effects get weaker. 

Other relevant work include a method based on Bayesian reduced rank regression recently presented in
\citet{foygel2012}, but it does not scale to the dimensionalities of our experiments either. Methods for
high-dimensional phenotypes have been proposed in the field of expression quantitative trait loci mapping
\cite{bottolo2011bayesian} but here the aim is to find associations (and stringently avoid false positives)
rather than the prediction of the phenotypes. Also functional assumptions not appropriate in our setup
\cite{wang2012integrative} have been used to constrain related learning problems.



\section{Model}

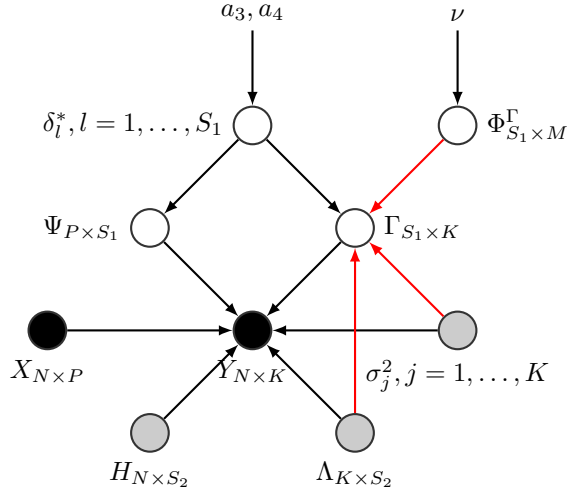
\begin{figure} 
\centering
\begin{tikzpicture}
\tikzstyle{main}=[circle, minimum size = 5mm, thick, draw =black!80, node distance = 14mm]
\tikzstyle{hyper2}=[draw=none, fill=none]
\tikzstyle{connect}=[-latex, thick]
\tikzstyle{box}=[rectangle, draw=black!100]
  \node[main, fill = black!100] (Y) [label=below:$Y_{N\times K}$] { };
  \node[main] (Psi) [above left=of Y, label=left:$\Psi_{P\times S_1}$] { };
  \node[main, fill = black!100] (X) [below left=of Psi, label=below:$X_{N\times P}$] { };
  \node[main] (Gamma) [above right=of Y, label=right:$\Gamma_{S_1 \times K}$] { };
  \node[main] (Gamma_shrink) [above right=of Gamma, label=right:{$\Phi_{S_1\times M}^{\Gamma}$}] { };
  \node[main] (delta_star) [above right=of Psi, label=left:${\delta_l^\ast, l=1,\ldots,S_1}$] { };
  \node[hyper2] (a3a4) [above=of delta_star] {$a_3,a_4$};
  \node[hyper2] (nu) [above=of Gamma_shrink] {$\nu$};
  \node[main, fill = black!20] (sigmas) [below right=of Gamma, label=below:{$\sigma_j^2, j=1,\ldots,K$}] { };
  \node[main, fill = black!20] (H) [below left=of Y, label=below:{$H_{N\times S_2}$}] { };
  \node[main, fill = black!20] (Lambda) [below right=of Y, label=below:{$\Lambda_{K\times S_2}$}] { };
  \path (X) edge [connect] (Y)
  	(Psi) edge [connect] (Y)
  	(Gamma) edge [connect] (Y)
  	(delta_star) edge [connect] (Psi)
  	(Gamma_shrink) edge [connect, red] (Gamma)
  	(delta_star) edge [connect] (Gamma)
		(a3a4) edge [connect] (delta_star)
		(sigmas) edge [connect] (Y)
		(H) edge [connect] (Y)
		(Lambda) edge [connect] (Y)
		(sigmas) edge [connect, red] (Gamma)
		(Lambda) edge [connect, red] (Gamma)
		(nu) edge [connect] (Gamma_shrink);
\end{tikzpicture}
\caption{Graphical representation of the model. The observed data are denoted by black circles, variables
related to the
reduced rank regression part of the model by white circles, and variables related to the noise model are
denoted by gray circles. Hyperparameters related to the noise model,
$\left\{\Phi^{\Lambda},a_{1},a_{2},\delta_{l},a_\sigma,b_\sigma\right\}$, are omitted for clarity. The
variable $\Phi_{S_1\times M}^{\Gamma}$ comprises the group sparsity parameters for the $M$ groups for the
different ranks, enforcing group-wise sparsity on the rows of $\Gamma_{S_1 \times K}$. Edges from $\sigma_j^2$
and $\Lambda_{K\times S_2}$ to $\Gamma_{S_1 \times K}$ facilitate the exploitation of the learned output
structure in the distribution of the regression coefficient matrix.}
\label{fig:dag}
\end{figure}

We propose to use the model%
\begin{equation}
Y=X\Psi\Gamma+H\Lambda^{T}+E,\label{eq:brr_model}%
\end{equation}
where $Y_{N\times K}$ contains the $K$-dimensional response variable from $N$
individuals, $X_{N\times P}$ contains the predictor variables, $\Psi_{P\times
S_{1}}$ and $\Gamma_{S_{1}\times K}$ represent a low-rank approximation for
the regression coefficient matrix $\Theta=\Psi\Gamma$, $H_{N\times S_{2}}$
contains unknown latent factors with the corresponding coefficient matrix
$\Lambda_{K\times S_{2}}$, and $E_{N\times K}=[e_{1},\ldots,e_{N}]^{T},$ with
$e_{i}\sim N(0,\Sigma),$ where $\Sigma=diag(\sigma_{1}^{2},\ldots,\sigma
_{K}^{2})$. Note that by integrating over the hidden factors $H$, the model becomes
equivalent to%
\[
y_{i}\sim N(\Theta^{T}x_{i},\Lambda\Lambda^{T}+\Sigma),\quad i=1,\ldots,N,
\]
which follows by assuming that the factors follow independent standard normal
distributions. 
Figure~\ref{fig:dag} displays graphically the structure of the model. In the figure, the edges from the noise
model ($\sigma_j^2$ and $\Lambda_{K\times S_2}$) to the regression coefficient parameter $\Gamma_{S_1 \times
K}$ represent the fact that a joint prior distribution for these parameters is specified, encoding the
assumption that effects on correlated phenotypes are likely to be similarly correlated.

Similarly to the Bayesian infinite sparse factor analysis model \cite{bhattacharya2011sparse}, we assume the
number of columns, $S_{2}$, in the weight matrix for the latent variables, $\Lambda$, to be infinite. This
way the problem of selecting a fixed rank is avoided. We exploit this idea further by letting also the rank
of the regression coefficient matrix, $S_{1}$, be infinite. The low-rank nature of the model is enforced by
shrinking the columns of $\Psi$ and $\Lambda$, and the rows of $\Gamma$, increasingly with the growing
column/row index, such that only the first columns/rows are influential in practice.

The priors for the parameters of the low-rank covariance inducing part of the model,
$H\Lambda+E$, where $\Lambda=[\lambda_{jh}]$, are set as in \citet{bhattacharya2011sparse}, as follows:%
\begin{gather}
\lambda_{jh}|\phi_{jh}^\Lambda,\tau_{h}\sim N\left(  0,\left(  \phi_{jh}^\Lambda\tau
_{h}\right)  ^{-1}\right), \hspace{1mm}\phi_{jh}^\Lambda\sim\text{Ga}(\nu/2,\nu/2),\nonumber\\ 
\tau_{h}=\prod_{l=1}^{h}\delta_{l}, \quad \delta_{1}\sim\text{Ga}(a_{1},1),\quad\delta_{l}\sim\text{Ga}(a_{2},1),\quad
l\geq2,\quad\nonumber\\
\sigma_{j}^{-2}\sim\text{Ga}(a_{\sigma},b_{\sigma}),\quad(j=1,\ldots
,K),\label{eq:Lambda_prior}%
\end{gather}
where $\tau_{h}$ is a global shrinkage parameter for the $h$th column of $\Lambda$ and
$\phi_{jh}^\Lambda$s are local shrinkage parameters for the elements in the $h$th
column, to provide additional flexibility to the normal distribution. It has been proven that $a_1>2$ and
$a_2>3$ is sufficient for the elements of $\Lambda\Lambda^T$ to have finite variance despite the infinite
number of columns in $\Lambda$ \cite{bhattacharya2011sparse}.

To facilitate the low-rank characterization of the regression coefficient matrix,
$\Theta=\Psi\Gamma$, we introduce a prior similar to ($\ref{eq:Lambda_prior}%
$). For the matrix $\Psi=\left[  \psi_{jh}\right]  $:
\begin{align}
\psi_{jh}|\tau_{h}^{\ast}  & \sim N\left(  0,\left(  \tau_{h}^{\ast}\right)
^{-1}\right)  ,\quad\tau_{h}^{\ast}=\prod_{l=1}^{h}\delta_{l}^{\ast
},\nonumber\\
\delta_{1}^{\ast}  & \sim\text{Ga}(a_{3},1),\quad\delta_{l}^{\ast}%
\sim\text{Ga}(a_{4},1),\quad l\geq2,\label{eq:Psi_prior}%
\end{align}
where the parameter $\tau_{h}^\ast$ acts as a global shrinkage parameter for the
$h$th column of $\Psi$.

In order to derive the group-sparsity inducing priors for $\Gamma,$ first let
$\mathcal{C=}\{C_{1},\ldots,C_{M}\}$ denote a pre-specified partitition of the
response variables such that $C_{i}, i=1,\ldots,M$, are disjoint non-empty subsets of the $K$
variables whose union includes all the variables. A multivariate normal distribution is placed independently
on each row of $\Gamma$, such that the covariance matrix consists of blocks corresponding to the pre-specified
groups. By using notation $\Gamma=[\gamma_{jh}]$ and $\gamma_{j\cdot}$ for the $j$th row of $\Gamma$, the
prior can be written as follows:
\begin{equation}
\gamma_{j\cdot}|\phi_{j\cdot}^{\Gamma},\tau_{j}^{\ast},\Lambda,\sigma_j^2\sim N_{K}\left(
0,\left(  \tau_{j}^{\ast}\right)  ^{-1}\Sigma_{j}^{\Gamma}\right)
,\label{eq:Gamma_prior}%
\end{equation}
where $\Sigma_{j}^{\Gamma}$ is a $K\times K$ covariance matrix. To denote a
submatrix of a given matrix corresponding to certain rows and columns, we
employ notation where the row and column indices are given in parenthesis
after the matrix symbol. For example $\Sigma_{j}^{\Gamma}(C_{1},C_{2})$
denotes the sub-matrix of the covariance matrix $\Sigma_{j}^{\Gamma}$
corresponding to rows from set $C_{1}$ and columns from $C_{2}$. We complete
our model formulation by specifying $\Sigma_{j}^{\Gamma}$ through
\begin{equation}
\Sigma_{j}^{\Gamma}(C_{m},C_{n})=\left\{
\begin{array}
[c]{c}%
\left(  \phi_{jm}^{\Gamma}\right)  ^{-1}\Sigma^{\ast}(C_{m},C_{n}%
),\quad\text{if }m=n, \text{and}\\
0,\quad\text{if }m\neq n.
\end{array}
\right.  \label{eq:Gamma_row_cov}%
\end{equation}
In equation (\ref{eq:Gamma_row_cov}), $\Sigma^{\ast}$ is the correlation matrix
between the response variables, obtained by normalizing the residual
covariance matrix $\Lambda\Lambda^{T}+\Sigma$; hence the conditioning on $\Lambda$ and $\sigma_{j}^2$ in Equation (\ref{eq:Gamma_prior}).
Furthermore, $\phi_{jm}^{\Gamma}$ is the shrinkage parameter for the
elements on the $j$th row of $\Gamma$ representing effects on the $m^{th}$ group $C_{m}$ of response
variables. 

Intuitively, equation (\ref{eq:Gamma_row_cov}) shows how the current estimate of the residual correlation
matrix for a pre-specified group of phenotypes enters the prior covariance of the regression coefficients
representing effects targeted to the same group of phenotypes, and this is the key point in the sharing of
information between the noise model and the regression model. For additional regularization, the group-wise
shrinkage parameters can enhance or suppress effects over the pre-specified groups of correlated variables.
From (\ref{eq:Gamma_row_cov}) we also see that the groups are assumed uncorrelated \textit{a priori} which
allows affordable computation.

Note that in (\ref{eq:Gamma_prior}), the parameters $\tau_{j}^{\ast}$ represent global shrinkage parameters
for the \textit{rows} of $\Gamma,$ as opposed to (\ref{eq:Lambda_prior}) and (\ref{eq:Psi_prior}) in which the
\textit{columns} were shrunk. Furthermore, the parameters $\tau_{h}^{\ast}$ and $\delta_{l}^{\ast}$ governing
the row and columnwise shrinkage of $\Gamma$ and $\Psi$ are shared between the priors of $\Psi$ and $\Gamma$
given in (\ref{eq:Psi_prior}) and (\ref{eq:Gamma_prior}). This means that the global shrinkage applied on each
column of $\Psi$ is equal to the shrinkage applied on the corresponding row of $\Gamma$, solving the
nonidentifiability resulting from the fact that the scales of $\Psi$ and $\Gamma$ can not be estimated
independently of each other.

\section{Computation}

For estimating the parameters of the model, we use the Gibbs sampling, updating the parameters one-by-one by
sampling them from their conditional posterior
probability distributions given the current values of all other parameters.
Here, the Gibbs sampling proceeds by alternatingly updating the reduced
rank regression part of the model and the noise model. For updating the
parameters related to the reduced rank regression part,
$\{\Psi,\Gamma,\Phi^{\Gamma},a_{3},a_{4},\delta_{l}^{\ast}\}$, we first compute
the residuals%
\[
Y^{\ast}=Y-H^{(i)}\Lambda^{T(i)},
\]
where $H^{(i)}$ and $\Lambda^{T(i)}$ are the current values of the corresponding variables. The standard Gibbs sampling updates for the Bayesian reduced rank regression from \cite{Geweke96} can be used with minor modifications to update the reduced rank regression parameters, given the residuals. Notice that the standard update can also be applied for $\Gamma$, once the prior distribution of $\Gamma$ that depends on the current estimate of the noise model parameters ($\Lambda, \sigma_j^2$) has been updated accordingly. 

To update the noise model parameters $\left\{ 
\Lambda,H,\Phi^{\Lambda},a_{1},a_{2},\delta_{l},\sigma_j^2\right\}$, we calculate the residuals
\[
Y^{\ast\ast}=Y-X\Psi^{(i)}\Gamma^{(i)},
\]
where $\Psi^{(i)}$ and $\Gamma^{(i)}$ are now the current values of the $\Psi$ and $\Gamma$ parameters. Given
the residuals $Y^{\ast\ast}$ the Gibbs sampling updates for the Bayesian infinite sparse factor analysis model
\cite{bhattacharya2011sparse} can be applied to the noise model parameters, which is essentially equivalent
to estimating a low-rank approximation for the
residual covariance matrix. We note that when updating the noise model, we should take into account not only
the current residuals from the $N$ samples, but also the $S_1$ rows from the $\Gamma$ matrix, which are
assumed to follow the same covariance pattern as the residuals save for the group-sparse structure.
However, in our experiments $N$ is very large, in the hundreds or thousands, compared the number of rows of
$\Gamma$ with non-negligible values. Therefore, taking $\Gamma$ into account when updating the noise model is
expected to have a negligible effect on the results and we employ an approximation that $\Gamma$ is not
considered 
when 
updating the noise model. In principle, deriving a Metropolis-Hastings MCMC update to account for $\Gamma$ would be straightforward, for example by generating proposals using the approximate Gibbs sampling, and accepting or rejecting the proposals based on the standard Metropolis-Hastings acceptance ratio. However, for simplicity, this approach is not considered here.

\section{Proofs of validity}

In this section we prove some characteristics of the infinite reduced rank regression model. Specifically, we show first that, in analogy to the infinite Bayesian factor analysis model \cite{bhattacharya2011sparse}, 
\begin{equation}
a_3>2 \quad \text{and} \quad a_4>3
\label{eq:sufficient}
\end{equation}
is sufficient for the prediction of any of the response variables to have finite variance under the prior distribution (Propositions 1 and 2). Second, we show that the underestimation of uncertainty (variance) resulting from using a finite rank approximation to the infinite reduced rank regression model decays exponentially with the rank of the approximation (Proposition 3).

For notational clarity, let $\Psi_{h}$ denote in the following the $h^{\text{th}}$ column of the $\Psi$
matrix. A prediction for the $i$th response variable is obtained by
\begin{align*}
\widetilde{y_{i}}  & =x^{T}\Theta_{i}\\
& =x^{T}\sum_{h=1}^{\infty}\Psi_{h}\gamma_{hi} ~~\text{.}
\end{align*}
According to a standard result,
\begin{equation}
\text{Var}(\widetilde{y_{i}})=\text{E}(\text{Var}(\widetilde{y_{i}}|\tau^{\ast}))+\text{Var(E}(\widetilde{y_{i}}|\tau^{\ast})),\label{prior_pred_var}%
\end{equation}
where $\tau^{\ast}=(\tau_{1}^{\ast},\tau_{2}^{\ast},\ldots)$. The first term is derived below in Proposition 1 under the conditions stated in Equation (\ref{eq:sufficient}). The second term is derived in Proposition 2. First, we prove the following Lemma:

\noindent\textbf{Lemma 1:}%

\begin{equation}
\text{Var}(x^{T}\Psi_{h}\gamma_{hi}|\tau_{h}^{\ast})=\frac{\nu}{\nu-2}\left(
\tau_{h}^{\ast}\right)  ^{-2}\sum_{j=1}^{P}\text{Var}(x_{j})
\label{eq:one_component}%
\end{equation}
%
A detailed proof is provided in the
supplementary material. 

\noindent\textbf{Proposition 1:} Suppose that $a_{3}>2$ and $a_{4}>3$. Then%

\begin{equation}
\text{Var}(\widetilde{y_{i}})=\frac{\nu}{\nu-2}\sum_{j=1}^{P}\text{Var}%
(x_{j})\frac{\Gamma(a_{3}-2)/\Gamma(a_{3})}{1-\Gamma(a_{4}-2)/\Gamma(a_{4})}.
\label{eq:exp_ptve}%
\end{equation}
\noindent\textbf{Proof of Proposition 1:}
\noindent Under the prior distribution the columns of $\Psi$ and the rows of
$\Gamma$ are conditionally independent given $\tau^{\ast}$. Therefore, we can
write%
\begin{align}
\text{Var}(\widetilde{y_{i}}|\tau^{\ast}) &  =\text{Var}(x^{T}\sum
_{h=1}^{\infty}\Psi_{h}\gamma_{hi}|\tau^{\ast})\nonumber\\
&  =\sum_{h=1}^{\infty}\text{Var}(x^{T}\Psi_{h}\gamma_{hi}|\tau^{\ast
}).\label{eq:inf_sum}%
\end{align}
Here we have used the fact that the variance of an infinite sum of independent
random variables is equal to the sum of the individual variances, as long as
the latter exists and \noindent the sum of the expectations of the variables
converges. Now let $\delta^{\ast}=(\delta_{1}^{\ast},\delta_{2}^{\ast}%
,\ldots).$ Substituting (\ref{eq:one_component}) and $\tau_{h}^{\ast}%
=\prod_{l=1}^{h}\delta_{l}^{\ast}$ into (\ref{eq:inf_sum}) yields
\begin{align}
\sum_{h=1}^{\infty}&\text{Var}(x^{T}\Psi_{h}\gamma_{hi}|\delta^{\ast}%
)= \nonumber\\ &\frac{\nu}{\nu-2}\sum_{j=1}^{P}\text{Var}(x_{j})\sum_{h=1}^{\infty}\left(
\delta_{1}^{\ast}\right)  ^{-2}\prod_{l=2}^{h}\left(  \delta_{l}^{\ast
}\right)  ^{-2} \label{eq:var_exp_all_comps} ~\text{.}%
\end{align}

\noindent We can now take the expectation of (\ref{eq:var_exp_all_comps}) over
the prior distribution of $\delta^{\ast}$, as follows:%
\begin{align}
\text{E}&\left[  \sum_{h=1}^{\infty}\text{Var}(x^{T}\Psi_{h}\gamma_{hi}%
|\delta^{\ast})\right] \nonumber\\
& =\frac{\nu}{\nu-2}\sum_{j=1}^{P}\text{Var}%
(x_{j})\sum_{h=1}^{\infty}\text{E}[\left(  \delta_{1}^{\ast}\right)
^{-2}]\prod_{l=2}^{h}\text{E}[\left(  \delta_{l}^{\ast}\right)  ^{-2}%
]\nonumber\\
&  =\frac{\nu}{\nu-2}\sum_{j=1}^{P}\text{Var}(x_{j})\sum_{h=1}^{\infty
}\text{E}[(\delta_{1}^{\ast})^{-2}]\text{E}[(\delta_{2}^{\ast})^{-2}%
]^{h-1}\nonumber\\
&  =\frac{\nu}{\nu-2}\sum_{j=1}^{P}\text{Var}(x_{j})\frac{\text{E}\left[
(\delta_{1}^{\ast})^{-2}\right]  }{1-\text{E}\left[  (\delta_{2}^{\ast}%
)^{-2}\right]  }\label{eq:exp_var} ~\text{.}
\end{align}

\noindent The first equality follows by changing the order of summation and
integration (owing to the Fubini's theorem and the fact that all $\delta
_{i}^{\ast}>0$). The second equality follows because $\delta_{2}^{\ast}%
,\delta_{3}^{\ast},\ldots$ are independent and identically distributed.
Finally, the third equality follows by assuming
\begin{equation}
\text{E}\left[  (\delta_{1}^{\ast})^{-2}\right]  <\infty\text{\quad and\quad
E}\left[  (\delta_{2}^{\ast})^{-2}\right]  <1,\label{eq:condition}%
\end{equation}
and using the standard result for the sum of a geometric series. Under our
assumption, namely that $a_{3}>2$ and $a_{4}>3$, the expectations are
available in a closed form as%
\begin{equation}
\text{E}\left[  (\delta_{1}^{\ast})^{-2}\right]  =\frac{\Gamma(a_{3}%
-2)}{\Gamma(a_{3})}\label{eq:c}%
\end{equation}
and%
\begin{equation}
\text{E}\left[  (\delta_{2}^{\ast})^{-2}\right]  =\frac{\Gamma(a_{4}%
-2)}{\Gamma(a_{4})}.\label{eq:d}%
\end{equation}
Furthermore, it is easy to check that under the assumption the condition
(\ref{eq:condition}) is satisfied. Substitution of (\ref{eq:c}) and
(\ref{eq:d}) into (\ref{eq:exp_var}) leads to the stated result. $\blacksquare$

\noindent\textbf{Proposition 2:} Under the assumptions of Proposition 1: 
\begin{equation}
\text{Var(E}(\widetilde{y_{i}}|\tau^{\ast}))=0 \text{.}
\end{equation}

\noindent\textbf{Sketch of Proof of Proposition 2:}
\noindent The result intuitively follows from the fact that $\Psi_h$ and $\gamma_{hi}$ are conditionally
independent, given $\tau^*$, and have zero mean. A formal treatment requires the use of Fubini's theorem. A
detailed proof is provided in the supplementary material. 

\noindent\textbf{Proposition 3:} Let $\widetilde{y_{i}}^{S_{1}}$denote the
prediction for the $i$th phenotype when using an approximation for $\Psi$ and
$\Gamma$ consisting of the first $S_{1}$ columns or rows only, respectively.
Then,
\[
\frac{\text{Var}(\widetilde{y_{i}})-\text{Var}(\widetilde{y_{i}}^{S_{1}}%
)}{\text{Var}(\widetilde{y_{i}})}=\left[  \frac{\Gamma(a_{4}-2)}{\Gamma
(a_{4})}\right]  ^{S_{1}},
\]
that is, the reduction in the variance of the prediction resulting from using
the approximation, relative to the infinite model, decays exponentially with
the rank of the approximation.

A detailed proof is provided in the supplementary material. 

\section{Experiments and results}
We evaluate the new model ('sharing BRRR') on a real data set by comparing it with a state-of-the-art sparse
multiple-output regression method Graph-guided Fused Lasso ('GFlasso') by \citet{Kim09}, 
Bayesian linear model ('BLM') in \citet{gelman2004bayesian}, 
Bayesian reduced rank regression with an infinite shrinkage prior for the weight matrix ('shrinkage BRRR')
that is the new model without group sparsity and sharing information between noise model and regression
model, 
Bayesian infinite reduced rank regression with group sparsity ('group sparse BRRR') corresponding to the new
model without sharing information between noise model and regression model, 
L2 regularized multi-task learning ('L2 MTL'),
elastic net-penalized multi-task learning ('L2/L1 MTL'),
sparse Bayesian factor regression ('Bayesian factor regression')
and a baseline method of predicting with target data mean.

The data set comprises genome-wide SNP data along with metabolomics measurements for a cohort of 4,702
individuals \cite{rantakallio1969groups,soininen2009high}. To compare the methods, SNP data from 125 genes
known to be associated with some of the metabolites \cite{teslovich2010biological,Genetics_Consortium13} were
used to generate 125 test cases, where the phenotype consisting of 96 metabolites was predicted using the SNPs
from a gene as predictors. For each gene, different training sets with 940, 2351 and 3762 individuals were
sampled and the remaining individuals were used as a test set. 

To reduce the dimensionality of the genotype data, 100 SNPs were selected from each gene as a pre-processing
step using classical canonical correlation (CCA) analysis. A CCA model was learnt to connect each SNP to all
the metabolites simultaneously and 100 SNPs with the highest canonical correlations were selected for the
analysis \cite{tang2012gene}. In preliminary experiments, the predictions were unchanged when the number of
SNPs was reduced in this way from 800 to 100, but the computational speed-up was considerable with all
methods. Dimensionality reduction was more needed by comparison methods. L2/L1 MTL was
evaluated without feature selection to compare against methods that do not require variable selection.

GFlasso provides a suitable comparison as it encourages sharing of information between correlated responses, as our model, but does that in the Lasso-type penalized regression framework. Shrinkage BRRR includes the noise model to explain away confounders but only relies on the low-rank approximation to share statistical strength, that is, the sharing of information between the noise model and the regression model and the group-sparsity of the effects are not implemented. Group sparse BRRR is identical to sharing BRRR except for not implementing information sharing between the noise model and the regression model. Thus it allows the evaluation of the effect of sharing information. The BLM serves as a single-task learning baseline.

For GFlasso, the regularization parameters were selected from the default grid using cross-validation. The
method has been developed for genomic data indicating the default values should be appropriate. For a fair
comparison, the pre-specified correlation network required by the GFlasso was constructed to match the
group-wise sparsity assumptions of the sharing BRRR: correlations between the responses within the same
pre-defined clusters \cite{inouye2012novel} were set to the empirical correlations, and to 0
otherwise. Hyperparameters for all BRRR models and the BLM were integrated over using the MCMC. For the
sharing BRRR, its simplified versions and Bayesian factor regression 80,000 MCMC samples were generated,
40,000 were discarded as burnin (after which convergence surely was reached although with sharing BRRR similar
predictions were already achieved with 10,000 samples), and the remaining samples thinned with a factor of 10
were used for prediction. The 
truncation point of the infinite rank BRRR model was estimated from the
data using the adaptive procedure described in \cite{bhattacharya2011sparse}. The regularization parameters for L1/L2 MTL were selected using 10-fold cross validation. Parameter $\alpha$ controlling for the balance between L1 and L2 regularization was evaluated on the grid 0, 0.1, $\ldots$, 0.9, 1.0.

The methods were compared by computing MSE for the predictions of the metabolites in the test data.
Statistical significances of the performance differences between
different methods were evaluated with Student's paired t-test: different methods' MSEs were compared on each
gene-metabolite pair under the assumption of unequal variances. However,
for every gene used as a test case, only some of the metabolites could be predicted using SNPs in the gene.
This resulted in the overall MSE values of all methods being worse than the simple baseline. For this reason,
we computed the MSE values also using only those metabolites that could be predicted more accurately than the
baseline with at least one of the methods. Below we report results calculated both using all metabolites as
well as using only the predictable ones. 

Table \ref{Table_tulokset_kaikki} presents the results using all test data and Table
\ref{Table_tulokset_ennustuvat} on the subset of predictable metabolites. In both cases
and with all training data sizes, sharing BRRR outperforms the other methods. On the predictable metabolites,
the difference to the other methods is statistically significant ($p < 0.05$) except for GFlasso with training
data sizes $N =$ 3762 and $N =$ 2351. On all test data, the difference between sharing BRRR's and other
methods' performances is statistically significant, except for GFlasso (with $N =$ 2351) and the simplified
versions of sharing BRRR on the smallest training data $N =$ 940.

\begin{table}[ht]
\caption{Average test data MSEs (above) and p-values (below in brackets) for comparison with sharing BRRR. Sharing BRRR outperforms all other methods with all training set sizes. Baseline method achieves test data MSE of 1.00. P-values are computed using Student's paired t-test when comparing sharing BRRR to the other methods.}

\vskip 0.15in
 \centering
\small

\begin{tabular}{cl@{\hspace{2mm}}l@{\hspace{2mm}}ll@{\hspace{2mm}}}

 & $N =$3762 & $N =$2351 & $N =$940 \\
\hline
\begin{tabular}{@{}c@{}}\small{sharing} \\ \small{BRRR} \end{tabular}  & \textbf{1.004833} & \textbf{1.003515} & \textbf{1.002441} \\
\begin{tabular}{@{}c@{}}\small{GFlasso} \\ \small{} \end{tabular} &
\begin{tabular}{@{}c@{}} 1.004847 \\\scriptsize{(4.7e-01)} \end{tabular} &
\begin{tabular}{@{}c@{}} 1.003621 \\\scriptsize{(2.9e-12)} \end{tabular} &
\begin{tabular}{@{}c@{}} 1.002581 \\\scriptsize{(1.1e-23)} \end{tabular} \\
\begin{tabular}{@{}c@{}}\small{group sparse} \\ \small{BRRR} \end{tabular} &
\begin{tabular}{@{}c@{}} 1.005070 \\\scriptsize{(4.7e-01)} \end{tabular} &
\begin{tabular}{@{}c@{}} 1.003587 \\\scriptsize{(6.2e-09)} \end{tabular} &
\begin{tabular}{@{}c@{}} 1.002437 \\\scriptsize{(6.5e-01)} \end{tabular} \\
\begin{tabular}{@{}c@{}}\small{shrinkage} \\ \small{BRRR} \end{tabular} &
\begin{tabular}{@{}c@{}} 1.005098 \\\scriptsize{(3.1e-50)} \end{tabular} &
\begin{tabular}{@{}c@{}} 1.003593 \\\scriptsize{(5.1e-10)} \end{tabular} &
\begin{tabular}{@{}c@{}} 1.002440 \\\scriptsize{(9.4e-01)} \end{tabular} \\
\begin{tabular}{@{}c@{}}\small{L2 MTL} \\ \small{} \end{tabular} &
\begin{tabular}{@{}c@{}} 1.005175 \\\scriptsize{(2.0e-36)} \end{tabular} &
\begin{tabular}{@{}c@{}} 1.003686 \\\scriptsize{(9.5e-17)} \end{tabular} &
\begin{tabular}{@{}c@{}} 1.002663 \\\scriptsize{(1.4e-46)} \end{tabular} \\
\begin{tabular}{@{}c@{}}\small{L1/L2 MTL} \\ \small{} \end{tabular} &
\begin{tabular}{@{}c@{}} 1.005176 \\\scriptsize{(1.9e-36)} \end{tabular} &
\begin{tabular}{@{}c@{}} 1.003677 \\\scriptsize{(2.7e-16)} \end{tabular} &
\begin{tabular}{@{}c@{}} 1.002663 \\\scriptsize{(1.4e-46)} \end{tabular} \\
\begin{tabular}{@{}c@{}}\small{BLM} \\ \small{} \end{tabular} &
\begin{tabular}{@{}c@{}} 1.006949 \\\scriptsize{($\sim$0)} \end{tabular} &
\begin{tabular}{@{}c@{}} 1.007886 \\\scriptsize{($\sim$0)} \end{tabular} &
\begin{tabular}{@{}c@{}} 1.024744 \\\scriptsize{($\sim$0)} \end{tabular} \\
\begin{tabular}{@{}c@{}}\small{Bayesian} \\ \small{factor regression} \end{tabular} &
\begin{tabular}{@{}c@{}} 1.029424 \\\scriptsize{($\sim$0)} \end{tabular} &
\begin{tabular}{@{}c@{}} 1.040187 \\\scriptsize{($\sim$0)} \end{tabular} &
\begin{tabular}{@{}c@{}} 1.089679 \\\scriptsize{($\sim$0)} \end{tabular} \\
\hline

\end{tabular}

\label{Table_tulokset_kaikki}
\vskip -0.1in
\end{table}

\begin{table}[ht]
 \centering
\small
\caption{Average test data MSEs over such gene-metabolite pairs for which at least 1 method outperforms the baseline method. Sharing BRRR outperforms all other methods with all training set sizes. Baseline method achieves test data MSE of 1.00.}
\label{Table_tulokset_ennustuvat}

\vskip 0.15in
\begin{tabular}{cl@{\hspace{2mm}}l@{\hspace{2mm}}ll@{\hspace{2mm}}}
 

& $N =$3762 & $N =$2351 & $N =$940 \\
\hline
\begin{tabular}{@{}c@{}}\small{weak effects} \\ \small{BRRR} \end{tabular}  & 
\textbf{0.97013} & \textbf{0.98482} & \textbf{0.99128} \\
\begin{tabular}{@{}c@{}}\small{GFlasso} \\ \small{} \end{tabular} &
\begin{tabular}{@{}c@{}} 0.97017 \\\scriptsize{(2.4e-01)} \end{tabular} &
\begin{tabular}{@{}c@{}} 0.98483 \\\scriptsize{(5.7e-01)} \end{tabular} &
\begin{tabular}{@{}c@{}} 0.99140 \\\scriptsize{(7.5e-08)} \end{tabular} \\
\begin{tabular}{@{}c@{}}\small{group shrinkage} \\ \small{BRRR} \end{tabular} &
\begin{tabular}{@{}c@{}} 0.97053 \\\scriptsize{(6.7e-40)} \end{tabular} &
\begin{tabular}{@{}c@{}} 0.98485 \\\scriptsize{(3.7e-02)} \end{tabular} &
\begin{tabular}{@{}c@{}} 0.99132 \\\scriptsize{(1.8e-02)} \end{tabular} \\
\begin{tabular}{@{}c@{}}\small{shrinkage} \\ \small{BRRR} \end{tabular} &
\begin{tabular}{@{}c@{}} 0.97057 \\\scriptsize{(1.1e-41)} \end{tabular} &
\begin{tabular}{@{}c@{}} 0.98486 \\\scriptsize{(1.4e-02)} \end{tabular} &
\begin{tabular}{@{}c@{}} 0.99132 \\\scriptsize{(1.6e-02)} \end{tabular} \\
\begin{tabular}{@{}c@{}}\small{L2 MTL} \\ \small{} \end{tabular} &
\begin{tabular}{@{}c@{}} 0.97082 \\\scriptsize{(2.3e-52)} \end{tabular} &
\begin{tabular}{@{}c@{}} 0.98513 \\\scriptsize{(2.5e-26)} \end{tabular} &
\begin{tabular}{@{}c@{}} 0.99168 \\\scriptsize{(2.1e-97)} \end{tabular} \\
\begin{tabular}{@{}c@{}}\small{L1/L2 MTL} \\ \small{} \end{tabular} &
\begin{tabular}{@{}c@{}} 0.97081 \\\scriptsize{(1.3e-52)} \end{tabular} &
\begin{tabular}{@{}c@{}} 0.98512 \\\scriptsize{(8.2e-27)} \end{tabular} &
\begin{tabular}{@{}c@{}} 0.99168 \\\scriptsize{(2.1e-97)} \end{tabular} \\
\begin{tabular}{@{}c@{}}\small{BLM} \\ \small{} \end{tabular} &
\begin{tabular}{@{}c@{}} 0.97215 \\\scriptsize{(2.8e-281)} \end{tabular} &
\begin{tabular}{@{}c@{}} 0.98917 \\\scriptsize{($\sim$0)} \end{tabular} &
\begin{tabular}{@{}c@{}} 1.01427 \\\scriptsize{($\sim$0)} \end{tabular} \\
\begin{tabular}{@{}c@{}}\small{Bayesian} \\ \small{factor regression} \end{tabular} &
\begin{tabular}{@{}c@{}} 0.99496 \\\scriptsize{($\sim$0)} \end{tabular} &
\begin{tabular}{@{}c@{}} 1.02216 \\\scriptsize{($\sim$0)} \end{tabular} &
\begin{tabular}{@{}c@{}} 1.08210 \\\scriptsize{($\sim$0)} \end{tabular} \\

\hline 

\end{tabular}

\vskip -0.1in

\end{table}

The difference between sharing BRRR and group sparse BRRR is statistically significant demonstrating the
importance of sharing information between the noise model and the regression model. With the larger training
set sizes the performance of GFlasso is almost as good as that of our model, which is not surprising as both
methods encourage effects on correlated phenotypes to be similar. However, this is accomplished in the two
methods based on completely different statistical principles, and it is assuring to see the performances of
the two methods converging towards each other when the training set sizes increase.

\begin{figure}[ht!]
\centering
\includegraphics[width=0.47\textwidth]{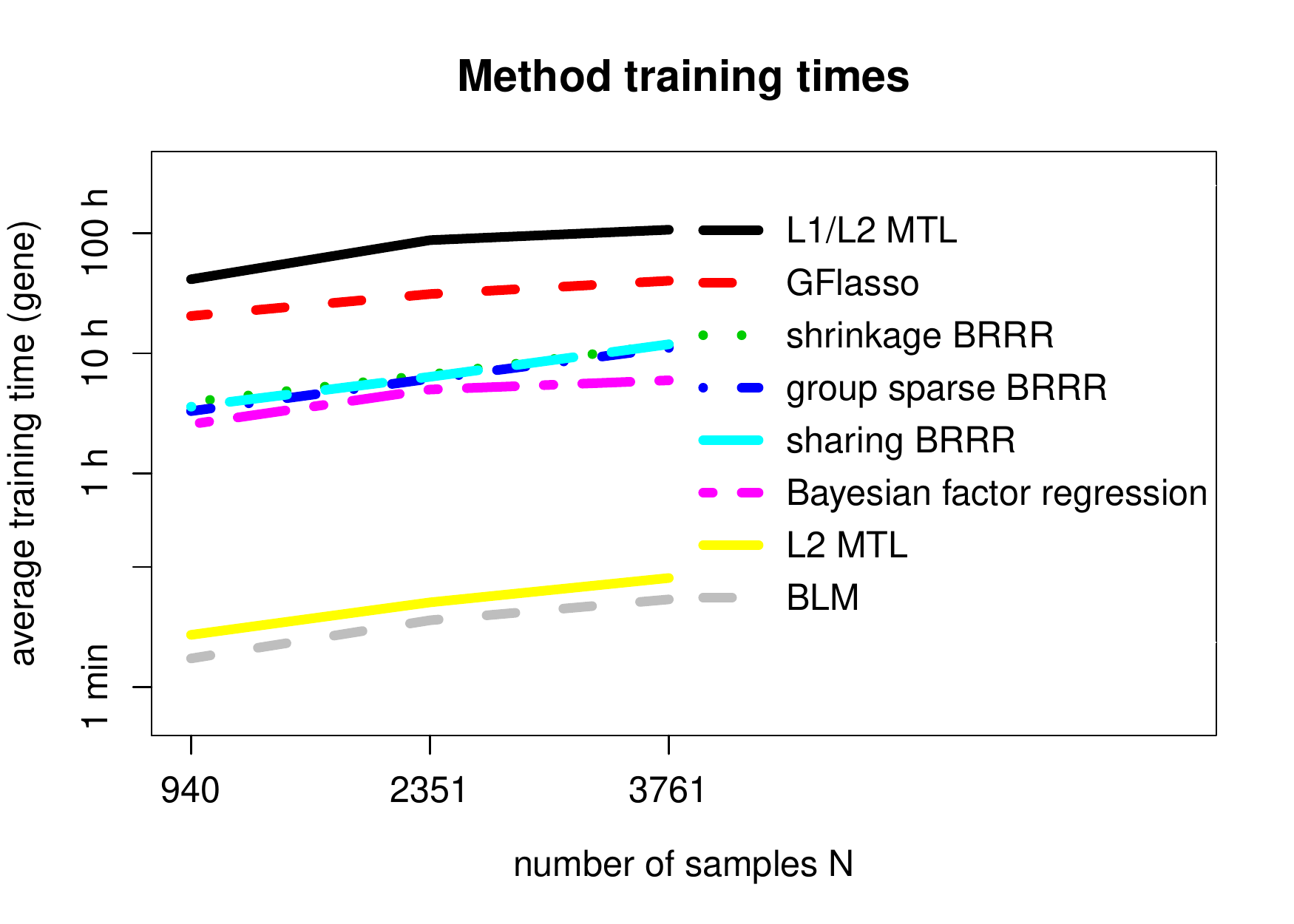} \\

\caption{Training times for all methods as a function of the number of training data.}
\label{Fig_ajoajat}
\end{figure}

Finally, we emphasize that the sharing BRRR was here treated in a fully Bayesian manner and no cross
validation to select hyperparameters was performed. It is our future plan to investigate whether the
prediction accuracy could be improved even further by optimizing the model parameters in this way. One of the
advantages of the fully Bayesian treatment is apparent in Figure \ref{Fig_ajoajat}: the training times for the
BRRR methods are approximately one order of magnitude smaller than those of GFlasso, which was the closest
method in terms of prediction performance.

\section{Discussion}

In this work we focused on the problem of predicting weak effects when data are abundant. This problem has
previously received only marginal attention, despite that fact that such data sets are rapidly accumulating in
many application fields. We concentrated on prediction (rather than finding associations); this is needed in
plant breeding and the need is emerging also in medical applications. We have shown that optimal predictive
performance with weak effects can be attained by combining techniques from the small $n$, large $p$ setup with
data sets comprising thousands of samples.

Our method incorporates the conceptually new feature of sharing information between the regression model and
the noise model in the Bayesian framework. In our experiments, this was the key aspect for being able to
predict any better than the baseline with some of the test data sets (i.e. genes) as otherwise the model would
have been too flexible to learn the weak effects. The benefit was clearly visible in the significant
improvement in predictive performance with all training set sizes compared to the model which did not include
the information sharing aspect, but was otherwise the same.

The new method reduces the effective number of parameters also by integrating current state-of-the-art machine
learning principles of rank reduction, group sparsity, nonparameteric shrinkage priors, and an intelligent
noise model in the multiple-output prediction setup. The concepts of group sparsity and infinite shrinkage
priors have previously not been introduced in the context of Bayesian reduced rank regression. Proofs
establishing the soundness of the infinite shrinkage prior and convergence of the truncation error were
provided. Our method allows a fully Bayesian treatment with realistic sample sizes (80,000 MCMC samples with a
training data sample size $N$ of 3761 in 10 hours). In this realistic-sized setup, the new method outperformed
the comparison methods when predicting weak effects in high-dimensional data.

\section*{Acknowledgments}

This work was financially supported by the Academy of Finland (grant number 251170 to the Finnish Centre of Excellence in Computational Inference Research COIN; grant number 259272 to PM; grant number 257654 to MP).
NFBC1966 received financial support from the Academy of Finland (project grants 104781, 120315, 129269, 1114194, 24300796, Center of Excellence in Complex Disease Genetics and SALVE), University Hospital Oulu, Biocenter, University of Oulu, Finland (75617), NHLBI grant 5R01HL087679-02 through the STAMPEED program (1RL1MH083268-01), NIH/NIMH (5R01MH63706:02), ENGAGE project and grant agreement  HEALTH-F4-2007-201413, EU FP7 EurHEALTHAgeing -277849 and the Medical Research Council, UK (G0500539, G0600705, G1002319,  PrevMetSyn/SALVE).
The DNA extractions, sample quality controls, biobank up-keeping and aliquotting was performed in the National Public Health Institute, Biomedicum Helsinki, Finland and supported financially by the Academy of Finland and Biocentrum Helsinki. We thank the late Professor Paula Rantakallio (launch of NFBC1966), and Ms Outi Tornwall and Ms Minttu Jussila (DNA biobanking). The authors would like to acknowledge the contribution of the late Academian of Science Leena Peltonen. 

\bibliographystyle{icml2014}
\bibliography{bib_EAS.bib}

\end{document}